\documentclass[11pt,a4paper]{article}
\usepackage[T1]{fontenc}
\usepackage[utf8]{inputenc}
\usepackage[a4paper,margin=2.5cm]{geometry}
\usepackage{microtype}
\usepackage{lmodern}
\usepackage{graphicx}
\graphicspath{{./}} 
\usepackage{booktabs}
\usepackage{array}
\usepackage[ruled,vlined]{algorithm2e}
\usepackage{soul}
\usepackage{xcolor}
\usepackage{tabularx}
\usepackage{longtable}
\usepackage{caption}
\usepackage{subcaption}
\usepackage{float}
\usepackage{placeins}
\usepackage{multirow}
\usepackage{booktabs}
\usepackage{tikz}
\usetikzlibrary{shapes.geometric, arrows.meta, 
                positioning, fit, backgrounds}
\usepackage{amsmath, amssymb}
\usepackage{enumitem}

\usepackage[hidelinks]{hyperref}
\usepackage[nameinlink,noabbrev]{cleveref} 
\usepackage{bookmark}

\usepackage[dvipsnames]{xcolor}
\usepackage{todonotes}

\title{Quality-preserving Model for Electronics Production Quality Tests Reduction}
\author{Noufa Haneefa\(^{1,*}\), Teddy Lazebnik\(^{1,2,x}\), Einav Peretz-Andersson\(^{1,x}\)\\\\ \(^1\) Department of Computing, Jonkoping University, Jonkoping, Sweden \\ \(^2\) Department of Information Science,  University of Haifa, Haifa, Israel \\ \(^x\) These authors contributed equally. \\ \(^*\) Corresponding author: \url{hano24vi@student.ju.se}}
\date{} 

\begin{document}

\maketitle

\begin{abstract}
Manufacturing test flows in high-volume electronics production are typically fixed during product development and executed unchanged on every unit, even as failure patterns and process conditions evolve. This protects quality, but it also imposes unnecessary test cost, while existing data-driven methods mostly optimize static test subsets and neither adapt online to changing defect distributions nor explicitly control escape risk. In this study, we present an adaptive test-selection framework that combines offline minimum-cost diagnostic subset construction using greedy set cover with an online Thompson-sampling multi-armed bandit that switches between full and reduced test plans using a rolling process-stability signal. We evaluate the framework on two printed circuit board assembly stages—Functional Circuit Test and End-of-Line test—covering 28,000 board runs. Offline analysis identified zero-escape reduced plans that cut test time by 18.78\% in Functional Circuit Test and 91.57\% in End-of-Line testing. Under temporal validation with real concept drift, static reduction produced 110 escaped defects in Functional Circuit Test and 8 in End-of-Line, whereas the adaptive policy reduced escapes to zero by reverting to fuller coverage when instability emerged in practice. These results show that online learning can preserve manufacturing quality while reducing test burden, offering a practical route to adaptive test planning across production domains, and offering both economic and logistics improvement for companies. 
\end{abstract}

\section{Introduction}
\label{sec:intro}
Ensuring that manufactured units conform to specified quality requirements before release is a fundamental objective of industrial production systems~\cite{scheme2009guide,duarte2025rethinking,grznar2025enhancing}. When defective products escape into the field, the resulting consequences extend beyond the direct costs of repair, replacement, and rework to include reputational damage, customer dissatisfaction, and loss of future business~\cite{Walston,tong2023understanding,bhowmick2023impact}. In high-volume manufacturing environments, these effects are amplified, making systematic quality assurance a critical component of operational performance~\cite{10952934,synnes2016bridging}. In this context, testing is one of the central mechanisms through which this assurance is achieved \cite{voorakaranam2002signature,milor1994minimizing}. Across modern production lines, products are subjected to structured sequences of diagnostic and functional checks, including electrical measurements, communication verification, firmware validation, and system-level performance evaluation, with the aim of identifying defective units before shipment~\cite{10417731,FOWLER20101,MAGNANINI2024349}.

Despite its importance, the design and execution of manufacturing test sequences remain challenging~\cite{Tahera,chowdhury2023design,fan2025embodied}. In most industrial settings, test flows are defined during product development on the basis of anticipated quality requirements and are then deployed as fixed procedures throughout production~\cite{6823865,Claudia}. Such staged verification strategies are widely used across domains. In semiconductor manufacturing, for example, early stages such as wafer sort assess die-level electrical integrity, whereas later stages such as package test verify the functionality of the packaged component before downstream processing~\cite{10812061}. Similarly, in biomedical additive manufacturing, early inspections evaluate dimensional accuracy and assembly integrity, while later stages address mechanical performance, sterilization, and regulatory compliance~\cite{11364033}. Although this fixed and comprehensive approach supports consistency and traceability, it does not account for the inherently dynamic nature of manufacturing environments, in which process variability, material changes, equipment drift, and evolving defect modes may alter the diagnostic value of individual test steps over time~\cite{Tran,11416281}. Consequently, test steps that were once highly informative may become redundant, while previously uninformative steps may gain diagnostic relevance as production conditions change~\cite{11329530}. At the same time, contemporary manufacturing systems continuously generate large volumes of process and test data that could support more adaptive decision-making~\cite{9353852}.

In parallel with these industrial developments, an increasing body of literature has examined the use of artificial intelligence (AI) and data-driven methods to improve manufacturing test efficiency \cite{plathottam2023review,ghelani2024ai,ghelani2024advanced}. Studies have shown that automation and learning-based approaches can improve repeatability and reduce operational cost in electronics testing environments~\cite{10560845,10568717}. Other studies have investigated the selection of diagnostically informative test items from historical data. For example, Anusuya et al.~\cite{10560845} and Agrawal et al.~\cite{10568717} demonstrated that AI-driven automation improves repeatability and reduces operational cost in electronics manufacturing test processes, yet neither addresses which test steps should be executed in the first place. Pan et al.~\cite{PAN2025102401} proposed an ensemble-learning framework that filters test items according to diagnostic contribution and reported a 34\% reduction in executed test items. However, the selected subset remains fixed after training, and the method does not address changes in defect distributions during production. 

Namely, meaningful reductions in testing effort are generally achieved under the assumption of stable operating conditions \cite{kim2022recent}. The problem of dynamically switching between full and reduced test plans during production, reverting to complete coverage when process conditions become unstable, and resuming reduced testing when stability is restored, remains largely unresolved. Equally important, prior approaches rarely quantify the defect escape risk associated with reduced testing or provide a mechanism for controlling that risk online. This limitation is particularly significant in manufacturing settings, where defect patterns may shift unexpectedly and where an initially safe reduced test plan may become insufficient when new failure modes emerge.

To this end, in this study, we address this gap by investigating whether historical production data can be used not only to identify a minimum-cost subset of diagnostically relevant test steps, but also to support an adaptive online policy that selects between full and reduced test plans on a per-unit basis. Specifically, we examine whether complete defect coverage can be preserved with a reduced subset derived from historical data, how the trade-off between test-time reduction and tolerated escape risk can be characterized, and whether an adaptive data-driven strategy can maintain quality performance under changing production conditions without offline retraining or manual reconfiguration. Formally, we propose an adaptive test-selection framework that combines offline subset optimization with online sequential decision-making. In the offline stage, a greedy set cover procedure is used to identify a minimum-cost diagnostic subset from historical production records~\cite{Adamo}. In the online stage, a Thompson Sampling-based multi-armed bandit (MAB) model~\cite{russo2018tutorial} dynamically selects between the reduced and full test plans for each incoming unit using real-time production evidence. Notably, previous studies demonstrated the usefulness of MAB-based models for test reduction but focused on software-testing only \cite{Lima}. This two-stage formulation is intended to reduce testing effort during stable operating periods while preserving full coverage when instability is detected. The proposed model is designed for production environments in which test steps are executed sequentially, outcomes are measurable, and quality performance can be assessed in terms of defect detection.

In order to evaluate the proposed model's performance, we explore its performance on Printed Circuit Board Assembly (PCBA) manufacturing conducted in collaboration with a large-size industrial electronics manufacturing company. PCBA is a representative and demanding application domain characterized by high production volume, strict quality requirements, and multi-stage diagnostic testing. We include two sequential stages, Functional Circuit Test (FCT) and End-of-Line (EOL) testing, comprising 172 and 55 individual test steps, respectively. Historical data from more than 28,000 board runs are used to analyze defect patterns and construct reduced test plans, while temporally separated validation data are used to assess performance under real concept drift conditions. 

The remainder of this paper is structured as follows. Section ~\ref{sec:related_work} briefly introduces manufacturing test optimization, data-driven test selection methods, and MAB algorithms for sequential decision-making. Section~\ref{sec:problem} formalizes the problem and outlines the proposed model. Section~\ref{sec:experiments} describes the experimental setup, followed by the obtained results. Finally, Section~\ref{sec:discussion} discusses the findings, limitations, and directions for future work.

\section{Related Work}
\label{sec:related_work}
In this section, we cover three core components of the proposed framework. First, the structure and economics of manufacturing test sequences are presented, establishing the context in which test optimization operates. Next, data-driven approaches to test subset selection and adaptive testing are outlined, with their limitations. Finally, we briefly present the MAB algorithm and its applications to sequential decision problems in manufacturing and quality control contexts, providing the theoretical foundation for the proposed model.

\subsection{Manufacturing test sequences}
\label{sec:test_flow}
Testing is a critical stage in the manufacturing process, ensuring that products meet design specifications and are free from quality defects. The problem of determining which tests should be executed and in what sequence, while minimizing total test time without compromising defect detection coverage, is a fundamental challenge in high-volume production environments~\cite{4395096,Wang2025Optimal,Yil2016Ada}. For instance, Scheffler et al.~\cite{namenamename} demonstrate that the relationship between test effort and quality exhibits diminishing returns, such that additional testing yields progressively smaller improvements in defect reduction as fault coverage approaches saturation. In a similar manner, Walston et al.~\cite{Walston} situate quality-related costs within the broader cost of quality framework and highlight that poor quality can impose financial consequences beyond direct rework and scrap costs, including lost sales, lost profits, reputational damage, and loss of repeat business. 

 An important implication of this cost-coverage trade-off is that the diagnostic contribution of individual test steps becomes highly relevant ~\cite{PAN2025102401,9801935}. Prior research suggests that diagnostic value within a test sequence is often unevenly distributed, such that some steps contribute substantially more to defect detection than others, while other steps add limited or redundant value \cite{Ferhani2008How,Benner}. To this end, Hamrol et al.~\cite{Hamrol} proposed a value-based framework for evaluating quality inspection in multi-stage manufacturing processes. In their model, inspection effectiveness is assessed in terms of added value, defined as the difference between quality costs with and without inspection. Their results indicate that inspection at a given stage should be abandoned or improved if it does not generate sufficient value. This reinforces the broader argument that not all test or inspection steps contribute equally to manufacturing quality outcomes. In addition, Liang and Zheng~\cite{9856920} present an industrial case study in semiconductor testing showing that optimizing the data extraction method within the test program can significantly reduce test time without altering the functional intent of the test sequence. Also, Iaria et al.~\cite{10812061} demonstrate this clearly in large-scale automotive SoC production. Using a weighted fault coverage metric applied to 80 million faults across a 40nm device, they show that the majority of defects are detected by a small fraction of available test patterns, whereas the remaining patterns contribute no additional diagnostic value, even after accounting for non-uniform defect distributions across the die.

\begin{figure}[htbp]
\centering
\includegraphics[width=0.9\textwidth]
{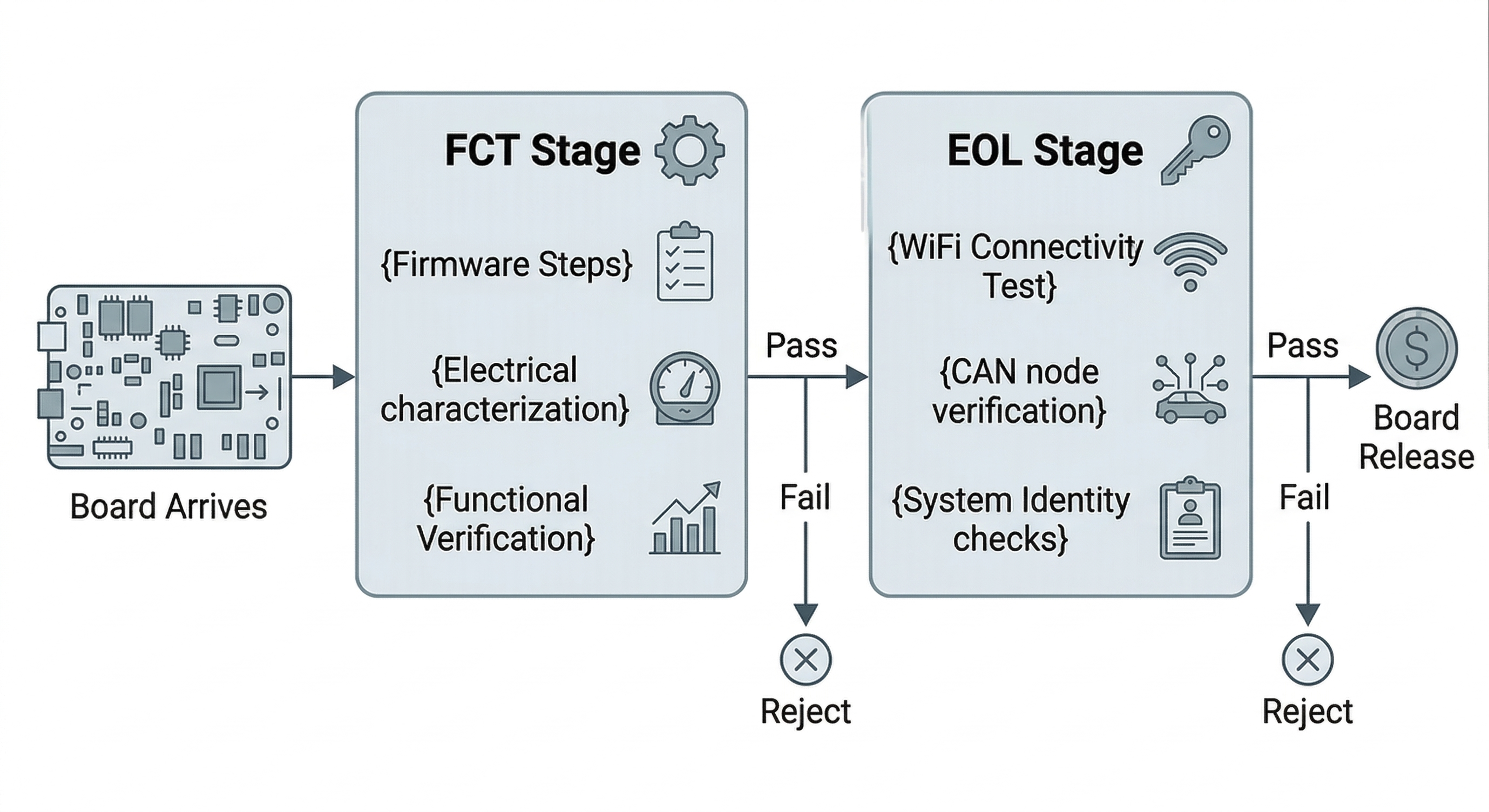}
\caption{PCBA manufacturing test flow}. 
\label{fig:pcba_flow}
\end{figure}

In the specific context of PCBA manufacturing, even a small defect in a single component can affect the functionality of the entire board \cite{Petkov}. To detect such defects, PCBA production employs multiple sequential test stages \cite{10270749}. In this process, the PCBA stage includes mandatory firmware programming steps that must be executed on every unit, like electrical characterization steps that verify power supply voltages and operating currents, and functional verification steps that confirm communication bus connectivity and system-level board identity \cite{Serban2014Uni,Kis2019ATS}. The End-of-Line stage subsequently validates complete wireless system performance before shipment. Figure~\ref{fig:pcba_flow} 
illustrates the overall test flow evaluated in this work.

\subsection{Data-driven test picking models}
\label{sec:data_driven}
Several studies have proposed data-driven approaches to reduce test time by identifying which test steps provide meaningful diagnostic information and which may be safely omitted by \cite{Ahsan2020Dev,KAUSIK2025100393}. These approaches vary in how test reduction is applied, with some identifying a fixed reduced subset is identified offline and applied uniformly to all units, others using correlation rules used to customize the test sequence based on process measurements, and in some using adaptive policies evolve over time as production results change ~\cite{6563786,Huang2016Test}. Notably, many of these approaches rely on historical production data to develop the initial test-reduction logic before deployment, followed by limited adaptation during production~\cite{SONG202240}

Specifically, for high-yield integrated circuit products, Pan et al.~\cite{PAN2025102401} proposed an ensemble-learning-based adaptive testing framework that addresses the severe class imbalance typical of such settings, where defective units represent a small fraction of production. In their framework, data re-balancing, feature selection, and decision boundary adjustment are combined to minimize the number of executed tests while maintaining high classification performance and limiting test escapes. Furthermore, Saha et al.~\cite{11117076} proposed an adaptive testing framework for post-manufacturing testing of compute-in-memory CNN accelerators, in which test images are applied progressively and testing stops once the device can be classified as pass or fail with sufficient statistical confidence. A sequential estimation variant further improves efficiency by applying test images in an ordered sequence, achieving up to 4.6$\times$ speedup. 

These models emphasize the need for adaptive test plan selection to handle shifts in production over time. Indeed, the selection of adaptive test plans has also been explored at the production level \cite{Arslan2011Ada}. For example, Rodrigues et al.~\cite{6563786} proposed a multi-agent system for adaptive functional test plan customization in a real washing machine production line, in which quality and process data collected across production stages are correlated using an MPFQ model to adjust the sequence of functional tests for each appliance dynamically, thereby reducing functional test time through the elimination of unnecessary test steps. However, the adaptation mechanism relies on predefined correlation rules rather than learning from observed test outcomes, and it does not provide explicit risk-bounded guaranties on defect escape or cost-constrained test step reduction on a per-unit level. In general, these approaches remain connected to historical data or predefined decision rules, with limited capacity for continuous online learning from evolving production conditions.

\subsection{Multi-Armed Bandit Algorithms}
\label{sec:mab}
\begin{figure}
    \centering
    \includegraphics[width=0.99\linewidth]{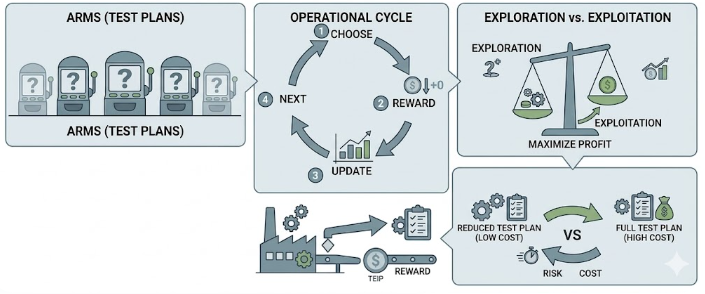}
    \caption{A schematic view of the multi-arm bandit algorithms and their production tests prediction.}
    \label{fig:mab}
\end{figure}

MAB algorithms provide an intuitive framework for repeated decision-making under uncertainty \cite{slivkins2019introduction,vermorel2005multi,mahajan2008multi}. The classical analogy is that of a gambler facing several slot machines, each with an unknown payoff distribution, who must decide which machine to play over time. In each round, the decision-maker must balance two competing objectives: exploiting the option that currently appears most promising and exploring alternatives that may prove superior as more evidence is accumulated. This exploration--exploitation trade-off makes the MAB framework particularly suitable for sequential industrial decisions in which actions must be selected repeatedly while their true value is only gradually revealed through observed outcomes~\cite{lazebnik2023data,lazebnik2024exploration,lazebnik2025publishing,russo2018tutorial,bouneffouf2020survey}. Figure \ref{fig:mab} provides a schematic view of the MAB algorithms and their production tests prediction.

Formally, a stochastic MAB problem is defined over a finite set of arms $\mathcal{A}=\{1,\dots,K\}$, where each arm represents a candidate decision and is associated with an unknown reward distribution $\nu_a$ with mean $\mu_a=\mathbb{E}[r \mid a]$~\cite{russo2018tutorial}. At each time step $t=1,\dots,T$, the agent selects an arm $a_t \in \mathcal{A}$ based on the accumulated history $\mathcal{H}_{t-1}=\{(a_s,r_s)\}_{s=1}^{t-1}$, observes a reward $r_t \sim \nu_{a_t}$, and updates its internal estimate of arm quality. The goal is to learn a policy $\pi$ that maximizes expected cumulative reward \(\mathbb{E}\!\left[\sum_{t=1}^{T} r_t\right]\), or equivalently minimizes cumulative regret relative to the best arm in hindsight
\begin{equation}
    R_T = T\mu^* - \mathbb{E}\!\left[\sum_{t=1}^{T} r_t\right]
    \label{eq:mab}
\end{equation}
where \(\mu^*=\max_{a\in\mathcal{A}}\mu_a\). In the context of adaptive manufacturing testing, the arms may correspond to alternative test plans, while the reward function can be formulated to reflect the operational objective, for example by rewarding test-time reduction and penalizing missed defect detections. Under this formulation, the exploration--exploitation trade-off becomes the problem of deciding when to apply a lower-cost reduced test plan and when to revert to a higher-cost but safer full test plan ~\cite{russo2018tutorial,bouneffouf2020survey}.

Several studies have demonstrated the usefulness of bandit methods in decision problems related to testing and quality control. In advanced manufacturing, Liu et al.~\cite{liu2023qualitytest} proposed a context-aware combinatorial bandit framework for quality testing in a B5G-enabled production environment. Their method uses contextual production information to decide which products should be tested under limited testing capacity, showing that bandit learning can support adaptive quality-test allocation in non-stationary industrial settings. In software testing, Lima and Vergilio~\cite{lima2022mab} introduced the COLEMAN approach, which uses a MAB to prioritize test cases in continuous integration environments based on historical failure information. Their results show that bandit-based policies can adapt to volatile test environments in which test cases are added or removed over time. These studies are closely related to the present work because they demonstrate the suitability of bandit methods for sequential testing decisions under uncertainty. However, the present study considers per-unit selection between alternative manufacturing test plans while explicitly managing defect escape risk.

\section{Task and Model Definition}
\label{sec:problem}
We define the task over a set of available test steps, each with an associated execution cost. The goal is to identify a subset of these steps that preserves defect detection coverage while reducing the total cost of testing each unit ~\cite{10812061}. Formally, let $\mathcal{T}=\{1,\dots,M\}$ denote the index set of available tests in a given stage, and let $c_i>0$ be the execution cost (e.g., mean time) of test $i\in\mathcal{T}$. The total cost of executing all test steps is \(c_{\text{full}} = \sum_{i \in \mathcal{T}} c_i\). This represents the per-unit test time under the current static test plan, which serves as the baseline cost against which all reductions are measured.

To capture the diagnostic behaviour of each step, consider $N$ historical units tested with the full flow. Let us define a binary outcome matrix $\mathbf{Y}\in\{0,1\}^{N\times M}$ with entries:
\begin{equation}
y_{u,i}=\begin{cases}
1 & \text{unit $u$ passed test $i$}\\
0 & \text{unit $u$ failed test $i$},
\end{cases}
\qquad u\in\{1,\dots,N\},\; i\in\mathcal{T}.
\label{eq:binary-outcomes}
\end{equation}
Each row of $Y$ represents the complete test outcome profile of one historical unit, and each column represents the pass/fail history of one test step across all units. Moreover, let $U_F=\{\,u:\exists i\in\mathcal{T}\;\text{s.t.}\;y_{u,i}=0\,\}$ be the set of historically failing units. In plain terms, $U_F$ contains every unit that failed at least one test step during the historical observation period, excluding equipment-induced failures. For any candidate subset $C\subseteq\mathcal{T}$, a failing unit $u\in U_F$ is \emph{detected} by $C$ if at least one executed test in $C$ fails historically \(D_C(u)=\max_{i\in C}\big(1-y_{u,i}\big)\in\{0,1\}\). Simply put, $D_C(u) = 1$ means subset $C$ would have caught defective unit $u$; $D_C(u) = 0$ means unit $u$ would have escaped detection under subset $C$. 

Furthermore, the empirical escape risk of $C$ is the fraction of historically failing units not detected by $C$, defined as \(\hat{R}(C)=\frac{1}{|U_F|}\sum_{u\in
U_F}\mathbf{1}\{D_C(u)=0\}\). A value of $\hat{R}(C) = 0$ means subset $C$ detects all historically failing units with zero escapes, while a value of $\hat{R}(C) = 1$ means all historically failing units would escape detection under subset $C$.

Based on this formalization, we defined three tasks: test value assessment, risk-bounded reduction, and dynamic test strategy. First, find the cheapest set of test steps that catches every defective unit observed in the training data:
\begin{align}
\min_{C\subseteq\mathcal{T}} \quad & \sum_{i\in C}c_i
\;\;\text{s.t.}\; \; \forall u\in U_F: D_C(u)=1.
\label{eq:rq1-con}
\end{align}
Second, find the minimum-cost subset under a tolerable empirical escape-risk threshold $\epsilon\in[0,1]$:
\begin{align}
\min_{C\subseteq\mathcal{T}} \quad & \sum_{i\in C}c_i
\;\;\text{s.t.}\;\;  \hat{R}(C)\le \epsilon.
\label{eq:rq2-con}
\end{align}
Third, let $C_{\mathrm{full}}=\mathcal{T}$ and $C_{\mathrm{red}}\subset\mathcal{T}$. A policy $\pi:[0,1]\to\{C_{\mathrm{full}},C_{\mathrm{red}}\}$ maps an observed process stability signal $\rho_t \in [0,1]$ to a test plan at each unit $t$. The goal is to minimize expected cost while bounding the policy-induced empirical escape risk by $\delta\in[0,1]$:
\begin{align}
\min_{\pi,\,C_{\mathrm{red}}}\quad & 
\mathbb{E}\!\left[\sum_{i\in \pi(\rho_t)} 
c_i\right] \;\; \text{s.t.} \;\; \hat{R}(\pi)\le \delta,
\label{eq:rq3-con}
\end{align}
where $\hat{R}(\pi)=\frac{1}{|U_F|}
\sum_{u\in U_F}
\mathbf{1}\{D_{\pi(\rho_t)}(u)=0\}$ is computed by replaying policy decisions over historical outcomes. To this end, unlike the first two tasks, which produce static subsets from historical data, this task requires a policy that adapts in real time to non-stationary production conditions, a sequential decision problem under uncertainty~\cite{9185782,lazebnik2026economical}.

Table ~\ref{tab:notation} summarizes the notations with their definitions.

\begin{table}[!ht]
\centering
\caption{Summary of the parameters used in the model 
with their notation and definition.}
\label{tab:notation}
\begin{tabular}{cl}
\hline \hline
\textbf{Symbol} & \textbf{Definition} \\
\hline \hline
$\mathcal{T} = \{1,\ldots,M\}$ & Index set of all available test steps \\$M$ & Total number of test steps \\$c_i$ & Execution cost (mean time in seconds) of test step $i$ \\$c_{\text{full}}$ & Total cost of executing all test steps \\$N$ & Number of historical units in training dataset \\$\mathbf{Y} \in \{0,1\}^{N \times M}$ & Binary outcome matrix \\$y_{u,i}$ & Outcome of unit $u$ on test step $i$ (1=pass, 0=fail) \\$U_F$ & Set of historically failing units \\$C \subseteq \mathcal{T}$ & A candidate subset of test steps \\$C_{\text{full}}$ & The complete set of all test steps $\mathcal{T}$ \\$C_{\text{red}}$ & The reduced subset \\$D_C(u)$ & Detection indicator for unit $u$ under subset $C$ \\$\hat{R}(C)$ & Empirical escape risk of subset $C$ \\$\epsilon$ & Tolerable escape risk threshold \\$\delta$ & Maximum allowed escape risk for MAB policy  \\$\pi$ & Test plan selection policy \\$\rho_t$ & Rolling pass rate at unit $t$ \\$w$ & Rolling window size \\$\tau$ & Process stability threshold \\$\beta$ & Instability sensitivity parameter \\$\kappa$ & Escape penalty in the reward function \\
$\alpha_a, \beta_a$ & Beta distribution parameters for arm $a$ \\
$\theta_a$ & Sampled value for arm $a$ \\$r_t$ & Reward signal at unit $t$ \\$a_t$ & Arm selected at unit $t$ \\ \hline \hline
\end{tabular}
\end{table}

With this formulation established, the proposed framework addresses the three research questions through two sequential phases. The offline phase solves the first two tasks using historical production data to construct the reduced test subset $C_{\text{red}}$ and characterize the cost-risk trade-off. In a complementary manner, the online phase solves the third task by deploying a Thompson Sampling MAB-based model that dynamically selects between $C_{\text{full}}$ and $C_{\text{red}}$ for each incoming unit based on a real-time process stability signal~\cite{russo2018tutorial, Nie,11248043}.

Formally, for the first task, we use the historical outcome matrix $\mathbf{Y}$ and cost vector $\mathbf{c}$ prior to deployment. These two stages construct the reduced subset $C_{\text{red}}$ that serves as the fixed alternative arm in the MAB agent. At each iteration, the greedy algorithm selects the test step that provides the greatest additional defect detection per unit of execution cost:

\begin{equation}
i^* = \arg\max_{i \in \mathcal{T} \setminus C} 
\frac{|\{u \in U_F : y_{u,i} = 0\} 
\setminus \text{Covered}(C)|}{c_i}
\label{eq:greedy}
\end{equation}

where $\{u \in U_F : y_{u,i} = 0\}$ is the set of failing units that step $i$ would detect, $\text{Covered}(C)$ is the set of failing units already detected by steps already in $C$, and the numerator counts only the \textit{newly} detected failing units that step $i$ would add. Dividing by $c_i$ ensure cheaper steps are preferred when they provide equivalent detection. Importantly, since the weighted set cover problem is NP-hard~\cite{Adamo}, an exact solution is intractable for large test suites. The greedy algorithm provides an approximation guarantee of $1 - 1/e \approx 63\%$ of the optimal solution in polynomial time, making it practical for industrial test suites with hundreds of steps ~\cite{Chvtal1979AGH,Adamo,10812061}. The output is the minimum-cost subset $C_{\text{red}} = C^*$ that satisfies $\hat{R}(C^*) = 0$.

Next, the second task extends the first task by sweeping the allowed escape threshold $\epsilon$ from $0$ to $|U_F|$, applying the greedy cover at each level to generate the full Pareto frontier of cost savings versus escape risk. Namely, the frontier $\mathcal{F}$ is initialized as empty. For each escape level $\varepsilon \in \{0, 1, \ldots, |U_F|\}$, Algorithm 1 is applied with the escape tolerance set 
to $\varepsilon / |U_F|$, 
producing a subset $C_\varepsilon$. 
The cost saving and empirical escape risk of $C_\varepsilon$ are then computed. Afterwards, a Pareto dominance check is performed inwhich the point is added to $\mathcal{F}$ only if no existing point in $\mathcal{F}$ simultaneously achieves both higher saving and lower escape risk. This ensures that $\mathcal{F}$ contains only non-dominated operating points, each representing the minimum-cost subset achievable at its corresponding escape tolerance level.

Finally, the last task is cast as a two-armed stochastic MAB. The two arms correspond to the full test plan $C_{\mathrm{full}}$ (arm 0) and the reduced test plan $C_{\mathrm{red}}$ (arm 1). At each production unit $t$, the agent selects one arm, executes the corresponding test plan, observes the outcome, and updates its belief about the value of each arm~\cite{9185782}. Before each arm selection decision, the agent computes a rolling pass rate over the last $w$ units:

\begin{equation}
\rho_t = \frac{1}{w}\sum_{i=\max(1,\,t-w+1)}^{t}
\mathbf{1}[\text{outcome}(u_i) = \text{PASS}]
\label{eq:rolling-passrate}
\end{equation}

where $w$ is the window size and $\mathbf{1}[\text{outcome}(u_i) = \text{PASS}]$ equals 1 if unit $u_i$ passed all executed steps and 0 otherwise. The rolling pass rate $\rho_t$ serves as a proxy for process stability: a high value (close to 1) indicates stable production where the reduced plan is likely safe; a low value signals elevated defect risk where the full plan should be preferred. The process is classified as stable if $\rho_t \geq \tau$, where $\tau$ is a pre-specified stability threshold. When the process is unstable ($\rho_t < \tau$), the Thompson Sampling score for $C_{\mathrm{full}}$ is boosted by an instability penalty \(\tilde{\theta}_0 = \theta_0 + \beta \cdot 
(\tau - \rho_t)\), where $\theta_0$ is the sampled value for arm 0, $\tau - \rho_t > 0$ is the magnitude of the instability, and $\beta \geq 0$ is the instability sensitivity parameter. A larger $\beta$ makes the agent more conservative, reverting to $C_{\text{full}}$ more readily when instability is detected. When the process is stable ($\rho_t \geq \tau$), no boost is applied and $\tilde{\theta}_0 = \theta_0$. Of note, this reward signal is designed to align agent behaviour with the quality-cost objective ~\cite{Nie, Liu}. The reward values reflect a strict quality hierarchy where defect escapes are penalized most severely, cost savings on clean units are rewarded maximally, and defect detection under the reduced plan receives a partial reward to acknowledge maintained quality coverage at reduced cost. For a unit $u_t$ processed under arm $a_t \in \{0, 1\}$, the reward is defined as:
\begin{equation}
r_t = \begin{cases}
0.0  & \text{if } a_t = 0 \;\;
(C_{\mathrm{full}} \text{ selected}) \\
1.0  & \text{if } a_t = 1 \text{ and unit is 
defect-free} \\
0.5  & \text{if } a_t = 1 \text{ and defect 
detected by } C_{\mathrm{red}} \\
-\kappa & \text{if } a_t = 1 \text{ and defect 
escapes } C_{\mathrm{red}}
\end{cases},
\label{eq:reward}
\end{equation}
where $\kappa > 0$ is the escape penalty. Selecting $C_{\text{full}}$ yields a neutral reward of zero -- it is safe but costly. Selecting $C_{\text{red}}$ on a clean unit 
yields the maximum reward. A detected defect under $C_{\mathrm{red}}$ yields a partial reward, signalling that quality was maintained but risk was present. An escape yields a large negative reward, discouraging the agent from selecting $C_{\mathrm{red}}$ under high-risk conditions.

\begin{figure}[!ht]
\centering
\includegraphics[width=1\textwidth]{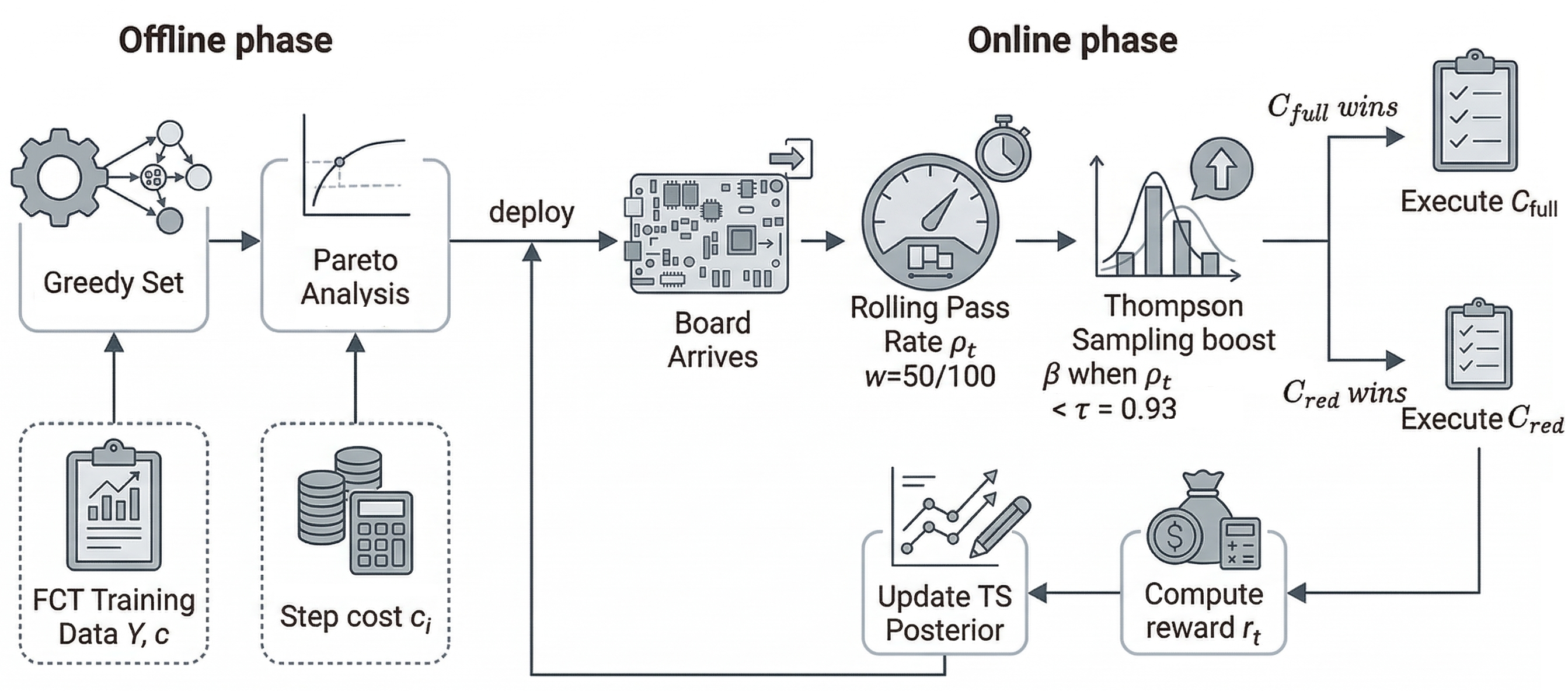}
\caption{Schematic flow of the proposed 
framework.}
\label{fig:framework}
\end{figure}

Figure~\ref{fig:framework} presents the schematic flow of the complete framework, from offline subset construction to online adaptive test execution, illustrating how the two phases interact to continuously adapt test plan selection to changing production conditions without requiring offline retraining ~\cite{russo2018tutorial, 
9185782}.

\section{Experiments}
\label{sec:experiments}
To evaluate the proposed model, we first outline a real-world experiment design and then present the obtained results of our analysis.

\subsection{Experiment design}
The experiments are conducted in collaboration with an industrial partner operating a high-volume electronics manufacturing line, where Printed Circuit Board Assemblies are produced and tested as part of the manufacturing process. In the current production process, every board undergoes two mandatory sequential test stages before integration into the final product: a Functional Circuit Test stage comprising 172 individual test steps that verify board-level electrical functionality and firmware integrity, and an End-of-Line stage comprising 55 test steps that validate complete wireless system performance. Any board failing either stage is rejected from the production flow. Under the current static test plan, the full sequence of steps is executed for every produced unit regardless of its production history or the current state of the production process. The mean full-sequence test time is 157.62 seconds per unit at FCT and 88.00 seconds per unit at EOL, representing a substantial proportion of the total per-unit production time. The production line operates under a strict quality threshold of $\delta = 8.5 \times 10^{-5}$ defect escapes per unit, which must be maintained at all times. However, executing the full test sequence for every unit regardless of current production conditions results in significant test time that can be reduced without compromising defect detection coverage under stable production conditions.

Formally, in this study, we use two datasets covering two structurally different test stages. Both datasets consist of raw test log files exported from the LabVIEW-based production test equipment, with one structured text file per board run recording the pass or fail outcome of each test step along with its execution time. The first dataset covers the Functional Circuit Test stage, comprising 172 distinct test steps executed across 7,618 board runs during the training period and 6,081 board runs during the validation period. Of the 332 board-level failures in the FCT training set, 302 are genuine defect events and 30 are abort-only runs attributed to test equipment errors rather than product defects, corresponding to a genuine defect rate of 3.96\% across the training period. The second dataset covers the End-of-Line test stage, comprising 55 distinct test steps executed across 10,633 board runs during training and 4,585 board runs during validation. Of the 416 board-level failures in the EOL training set, 318 are genuine defect events and 98 are abort-only runs, corresponding to a genuine defect rate of 2.99\%. The mean full-sequence test time is 157.62 seconds per board for FCT and 88.00 seconds per board for EOL.

Raw test log files were exported as structured text files, with one file per board run, and merged into a single flat CSV (Comma Separated Values) per dataset~\cite{Burg2019wra}. Column names were normalized to lowercase with underscores, and header-repeat rows- artefacts of the log export format were removed as part of standard data normalization and cleaning ~\cite{Sankpal2020normalization}. Each unique source filename was treated as a distinct unit identifier, as each file corresponds to one independent board run. A unique test step identifier was constructed by concatenating two hierarchical name fields, producing 172 unique step identifiers for FCT and 55 for EOL. Step execution times were cast to numeric values with missing entries filled as zero. Moreover, units whose board-level outcome was FAIL but whose individual step results contained no explicit FAIL, only ABORTs were classified as abort-only and excluded from the genuine failing unit set $U_F$. This filtering approach follows established practice in production test data cleaning, where equipment-induced failures are distinguished from genuine product defects prior to analysis. This exclusion removed 30 units from FCT and 98 units from EOL. The remaining 302 FCT and 318 EOL genuine failing units formed $U_F$ for all subsequent computations. In addition, a cost vector $\mathbf{c} \in \mathbb{R}^M$ of mean step execution times was derived by averaging step execution times across all runs for each step~\cite{9856920, 9801935}. The FCT data set comprises 7,618 training runs and 6,081 validation runs across 172 test steps, with 302 genuine defect events in training and 868 in validation a $3.6\times$ increase in defect rate. The EOL dataset comprises 10,633 training runs and 4,585 validation runs across 55 test steps, with 318 genuine defect events in training and 93 in validation. Both datasets are partitioned using a temporal split such that July-December 2025 forms the training set and January-February 2026 forms the held-out validation set.

In order to evaluate the proposed model's performance, four metrics are used and reported across both training and validation periods. Together they capture the two competing objectives of the framework: reducing test time and maintaining defect detection coverage. Table~\ref{tab:metrics} summarizes these four evaluation metrics in terms of notation, mathematical formalization, and motivation. 

\begin{table}[!ht]
\centering
\caption{Evaluation metrics used to assess the policy $\pi$.}
\label{tab:metrics}
\renewcommand{\arraystretch}{1.3}
\begin{tabularx}{\textwidth}{p{2.4cm} p{4.8cm} X}
\hline \hline
\textbf{Symbol} & \textbf{Equation} & \textbf{Explanation} \\
\hline \hline

$C_{\text{red}}$ selection rate (\(\mathrm{SelRate}(\pi)\))
& $\displaystyle 
\frac{1}{N}\sum_{t=1}^{N}\mathbf{1}[a_t=1]
$
& Measures the fraction of boards for which the MAB agent selected the reduced test plan rather than the full plan. Here, $a_t \in \{0,1\}$ is the arm selected for board $t$, and $N$ is the total number of boards processed. A value of $1.0$ means the agent always selected the reduced plan, while a value of $0$ means the full plan was always used. Higher values indicate greater exploitation of the cost-saving opportunity. \\

\hline

Cost saving (\%) (\(\mathrm{Saving}(\pi)\))
& $\displaystyle 
\left(1-\frac{\bar{c}_\pi}{c_{\text{full}}}\right)\times 100,
\qquad
\bar{c}_\pi=\frac{1}{N}\sum_{t=1}^{N}c_{a_t}
$
& Measures the percentage reduction in mean test time per board relative to executing the full sequence on every board. Here, $c_{a_t}$ is the cost of the plan selected for board $t$, and $c_{\text{full}}=\sum_{i\in\mathcal{T}} c_i$ is the cost of the full test sequence. A saving of $0\%$ corresponds to always running the full plan, while the maximum achievable saving corresponds to always running $C_{\text{red}}$. \\

\hline

Escaped units (\(\mathrm{Escaped}(\pi)\))
& $\displaystyle 
\sum_{u\in U_F}\mathbf{1}\{D_{\pi(y_u)}(u)=0\}
$
& Counts the absolute number of genuinely defective boards that were not detected by the executed test plan and would therefore have proceeded to shipment undetected. Here, $D_C(u)=\max_{i\in C}(1-y_{u,i})$. A value of zero means all defective boards were caught. Any non-zero value represents a direct quality failure: a defective board that reached the customer undetected. \\

\hline

Empirical escape risk (\(\hat{R}(\pi)\))
& $\displaystyle 
\frac{1}{|U_F|}\sum_{u\in U_F}\mathbf{1}\{D_{\pi(y_u)}(u)=0\}
$
& Expresses the same quantity as a fraction of the total defect population, making it comparable across datasets with different numbers of failing boards. A value of $\hat{R}(\pi)=0$ indicates complete defect coverage, meaning that all defective boards were detected. This metric is directly comparable to the quality threshold $\delta$, which requires $\hat{R}(\pi)\leq\delta$ at all times. \\

\hline \hline
\end{tabularx}
\end{table}

The proposed model is evaluated in three sequential stages. First, a greedy set cover algorithm~\cite{10812061} is applied to the training outcome matrix to identify the minimum-cost subset $C_{\text{red}} \subseteq \mathcal{T}$ that detects all genuine training failures with zero escape risk. At each iteration, the step with the highest ratio of newly detected failing units to execution cost is selected until all failing units are covered. Next, the Pareto frontier of cost saving versus escape risk is constructed by sweeping the allowed escape threshold from zero to $|U_F|$, generating one Pareto-optimal subset per unique cost-saving level~\cite{PAN2025102401}. For the FCT dataset, this is done using a dense sweep over all integer escape thresholds. For the EOL dataset, all $2^{14}$ possible subsets of diagnostic steps are enumerated exhaustively because the number of steps is small enough to allow such analysis. Finally, three MAB policies are trained online over the chronological unit stream: Thompson Sampling~\cite{russo2018tutorial}, Upper Confidence Bound (UCB)~\cite{9185782}, and Epsilon-Greedy~\cite{Nie}. Each policy is initialized with a conservative prior favoring $C_{\text{full}}$ ($\alpha_0=5, \beta_0=1$) and a uniform prior for $C_{\text{red}}$ ($\alpha_1=1, \beta_1=1$). The rolling pass-rate stability signal is computed over a sliding window of $w$ consecutive units and compared against threshold $\tau$ before each arm-selection decision. Thompson Sampling is then evaluated on the held-out validation stream under multiple instability sensitivity values ($\beta \in {10, 50}$ for FCT and $\beta \in {10, 50, 100}$ for EOL) to characterize the quality-cost trade-off under concept drift. 

\subsection{Results}
\label{sec:results}
Table~\ref{tab:combined_results} summarizes the results for both the FCT and EOL datasets across training and validation cohorts. In both datasets, all MAB methods achieve zero escapes during training while approaching the savings of the static reduced plan. Under validation, however, the static $C_{\text{red}}$ policy fails due to concept drift, producing 110 escapes on FCT and 8 on EOL. In contrast, Thompson Sampling adapts by shifting back toward the full plan when instability is detected, reducing escapes to zero at $\beta=50$ for FCT and $\beta=100$ for EOL. This shows that dynamic test selection preserves most of the cost benefit of reduction during stable periods while remaining robust to unseen failure modes at deployment.

\begin{table*}[!ht]
\centering
\small
\setlength{\tabcolsep}{4pt}
\caption{Comparative results on the FCT and EOL datasets across training and validation cohorts. $\beta$ is the instability sensitivity parameter, and $\hat{R}(\pi)$ denotes empirical escape risk.}
\label{tab:combined_results}
\vspace{4pt}
\begin{tabular}{lllc|cc|cc}
\hline \hline
\textbf{Test} & \textbf{Cohort} & \textbf{Algorithm} & \textbf{$\beta$} &
\textbf{$C_{\text{red}}$ (\%)} & \textbf{Saving (\%)} &
\textbf{Escaped} & \textbf{$\hat{R}(\pi)$} \\
\hline \hline

\multirow{9}{*}{FCT}
& \multirow{5}{*}{Training}
& Baseline ($C_{\text{full}}$) & --- & 0.0   & 0.00  & 0   & 0.00 \\
&  & Static $C_{\text{red}}$    & --- & 100.0 & 18.78 & 0   & 0.00 \\
&  & Epsilon-Greedy             & 10  & 89.6  & 16.83 & 0   & 0.00 \\
&  & UCB                        & 10  & 92.8  & 17.42 & 0   & 0.00 \\
&  & Thompson Sampling          & 10  & 93.8  & 17.61 & 0   & 0.00 \\
\cline{2-8}
& \multirow{4}{*}{Validation}
& Baseline ($C_{\text{full}}$) & --- & 0.0   & 0.00  & 0   & 0.00 \\
&  & Static $C_{\text{red}}$    & --- & 100.0 & 20.45 & 110 & 0.13 \\
&  & Thompson Sampling          & 10  & 16.1  & 3.30  & 8   & 0.01 \\
&  & Thompson Sampling          & 50  & 2.4   & 0.49  & 0   & 0.00 \\
\midrule

\multirow{10}{*}{EOL}
& \multirow{5}{*}{Training}
& Baseline ($C_{\text{full}}$) & --- & 0.0   & 0.00  & 0 & 0.00 \\
&  & Static $C_{\text{red}}$    & --- & 100.0 & 91.57 & 0 & 0.00 \\
&  & Epsilon-Greedy             & 10  & 87.0  & 79.68 & 0 & 0.00 \\
&  & UCB                        & 10  & 90.9  & 83.24 & 0 & 0.00 \\
&  & Thompson Sampling          & 10  & 91.5  & 83.77 & 0 & 0.00 \\
\cline{2-8}
& \multirow{5}{*}{Validation}
& Baseline ($C_{\text{full}}$) & --- & 0.0   & 0.00  & 0 & 0.00 \\
&  & Static $C_{\text{red}}$    & --- & 100.0 & 91.57 & 8 & 0.09 \\
&  & Thompson Sampling          & 10  & 98.2  & 89.95 & 1 & 0.01 \\
&  & Thompson Sampling          & 50  & 97.6  & 89.35 & 1 & 0.01 \\
&  & Thompson Sampling          & 100 & 95.2  & 87.17 & 0 & 0.00 \\
\hline \hline
\end{tabular}
\end{table*}

\begin{figure}[!ht]
\centering
\begin{minipage}{0.49\textwidth}
    \centering
    \includegraphics[width=\textwidth]{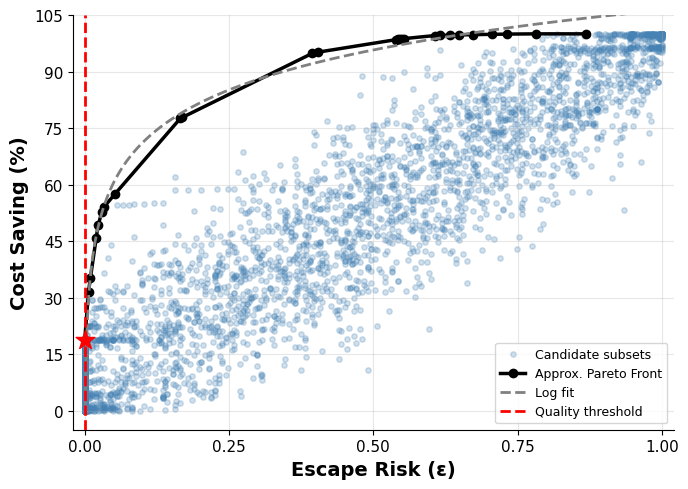}
    \subcaption{FCT Pareto front. }
\end{minipage}
\hfill
\begin{minipage}{0.49\textwidth}
    \centering
    \includegraphics[width=\textwidth]{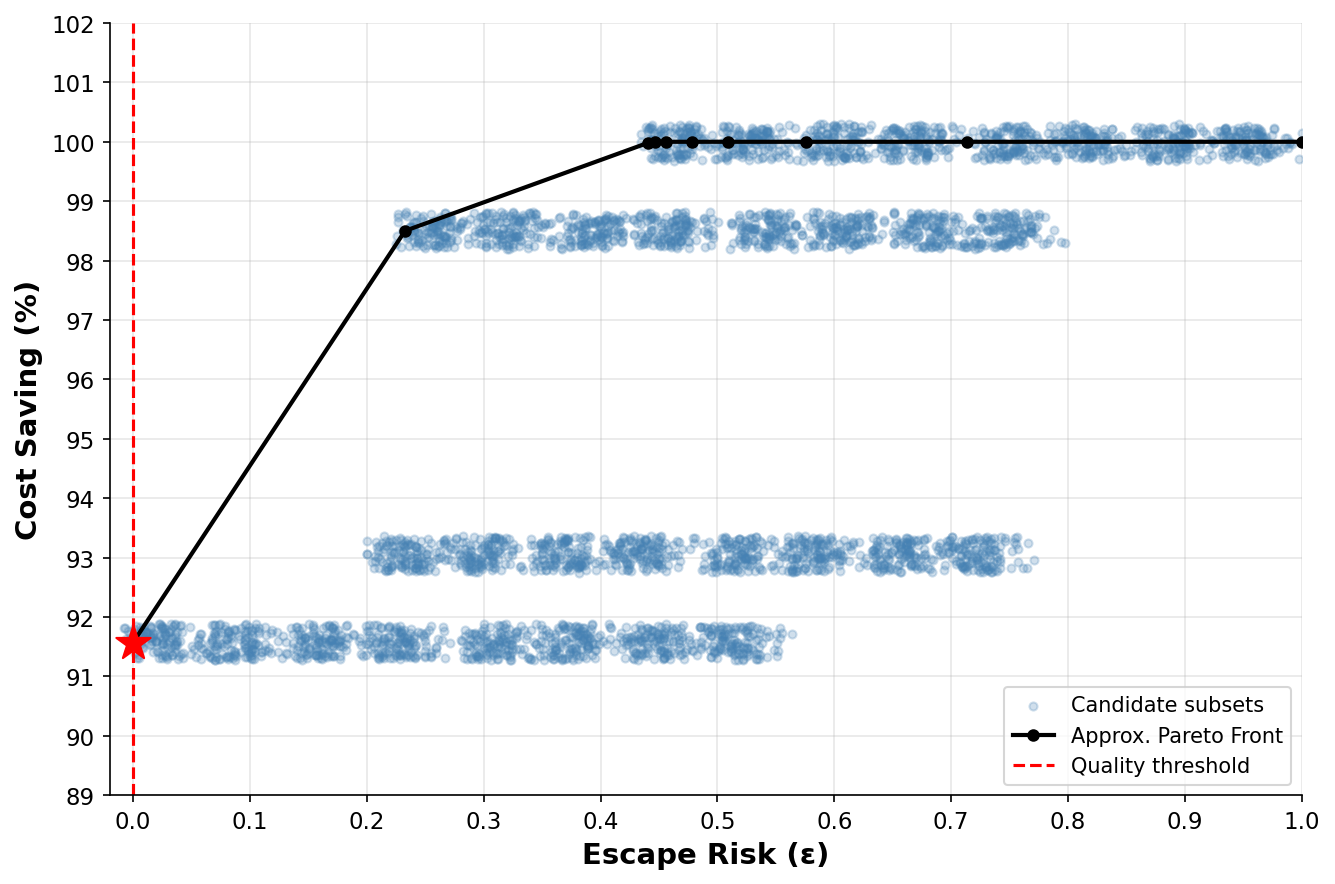}
    \subcaption{EOL Pareto front. }
\end{minipage}
\caption{Cost-risk Pareto fronts for the two test suites. The grey dotted line shows a logarithmic fit to the Pareto-optimal points, and the red star marks the zero-escape operating point.}
\label{fig:pareto_panels}
\end{figure}

Figure~\ref{fig:pareto_panels} shows the cost-risk Pareto fronts for both test suites. Panel (a) presents the FCT frontier, where the zero-escape operating point achieves 18.78\% saving using a 32-step subset that removes 140 redundant steps. Across all 13,699 FCT board runs, every step-level failure also produced a board-level failure, indicating that the rolling pass-rate signal captures all observed failure modes, including those outside $C_{\text{red}}$. The Pareto frontier follows a clear logarithmic trend, $\hat{S}(\varepsilon)=15.30\ln(\varepsilon)+106.34$ with $R^2=0.994$, showing strongly diminishing returns: most achievable savings are concentrated in the low-risk region near the operating point. Panel (b) presents the EOL frontier, obtained by exhaustively evaluating all $2^{14}$ subsets of diagnostic steps. Unlike FCT, the frontier exhibits visible discrete banding because only 14 diagnostic steps are available, so achievable saving levels are quantized. Despite this, the zero-escape operating point reaches 91.57\% saving, substantially higher than in FCT, because the 41 non-diagnostic EOL steps account for most of the total test time.

\begin{figure}[!ht]
\centering
\begin{minipage}{0.49\textwidth}
    \centering
    \includegraphics[width=\textwidth]{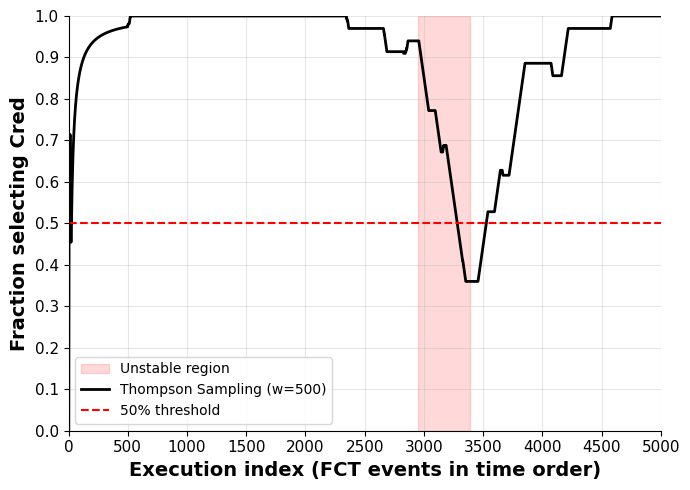}
    \subcaption{FCT training stream. The rolling fraction of units assigned to $C_{\text{red}}$ is shown for the Thompson Sampling agent ($w=300$). The detected instability region (units 2,950--3,400) is shaded in red.}
\end{minipage}
\hfill
\begin{minipage}{0.49\textwidth}
    \centering
    \includegraphics[width=\textwidth]{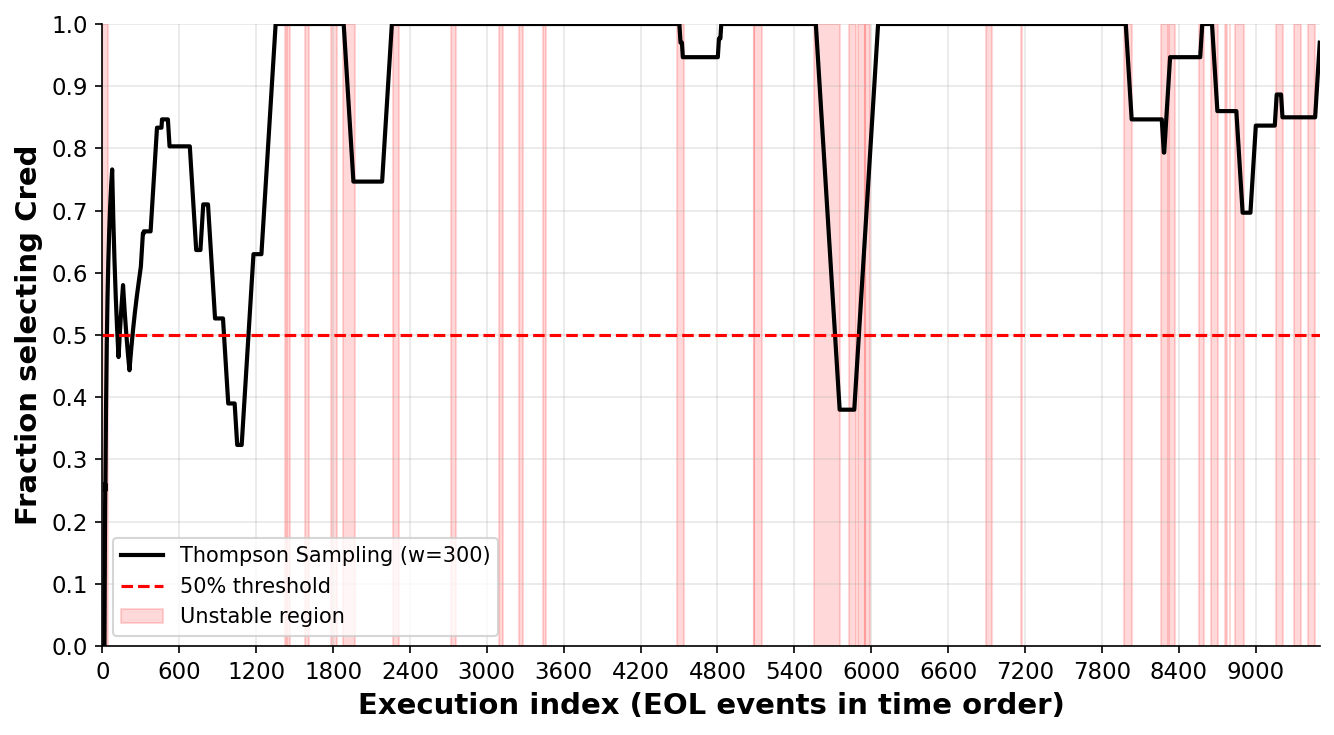}
    \subcaption{EOL training stream. The rolling fraction of units assigned to $C_{\text{red}}$ is shown for the Thompson Sampling agent. Multiple detected instability regions are shaded in red.}
\end{minipage}
\caption{Rolling $C_{\text{red}}$ selection fraction during training for the Thompson Sampling agent on the FCT and EOL datasets.}
\label{fig:armselection_panels}
\end{figure}

Figure~\ref{fig:armselection_panels} shows the rolling fraction of training units assigned to $C_{\text{red}}$ by the Thompson Sampling agent for both datasets. Panel (a) presents the FCT training stream over 7,618 units. The agent converges to about 95\% $C_{\text{red}}$ selection within the first 500 units, then drops sharply to nearly 35\% during the detected instability region before recovering, showing that the stability signal can temporarily override the learned cost preference when quality risk increases and that this response is reversible. Panel (b) presents the EOL training stream, where the agent converges to approximately 95\% $C_{\text{red}}$ selection by unit 1,500. Unlike FCT, which contains a single pronounced instability event, EOL exhibits multiple scattered instability regions across the training period, indicating a less stable production process. The agent responds to each episode by reducing $C_{\text{red}}$ selection and recovering afterward, demonstrating robust adaptive behavior under varying process conditions.

\begin{figure*}[!ht]
\centering
\begin{minipage}{0.49\textwidth}
    \centering
    \includegraphics[width=\textwidth]{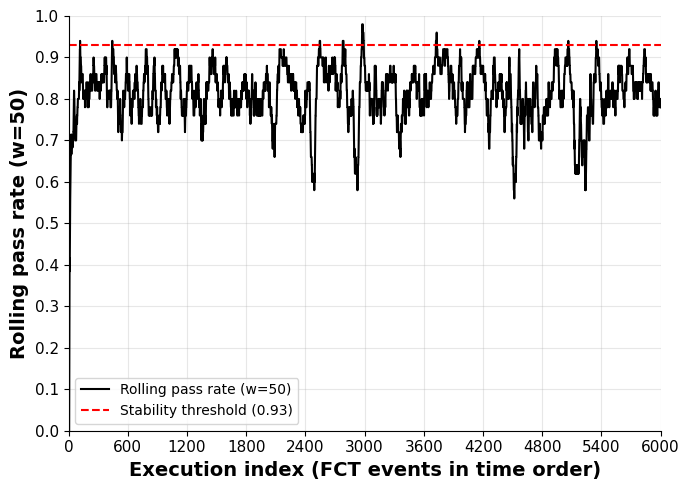}
    \subcaption{FCT validation rolling pass rate ($w=50$), which remains persistently below the stability threshold $\tau=0.93$.}
\end{minipage}
\hfill
\begin{minipage}{0.49\textwidth}
    \centering
    \includegraphics[width=\textwidth]{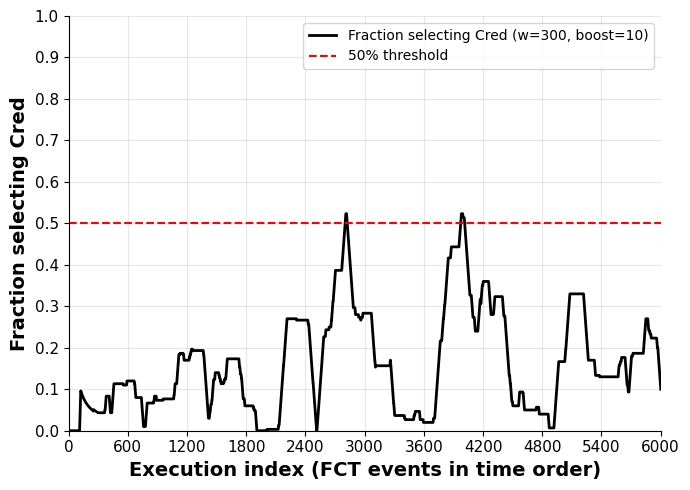}
    \subcaption{Rolling fraction of FCT validation units assigned to $C_{\text{red}}$ by the Thompson Sampling agent ($\beta=10$).}
\end{minipage}

\vspace{0.5em}

\begin{minipage}{0.49\textwidth}
    \centering
    \includegraphics[width=\textwidth]{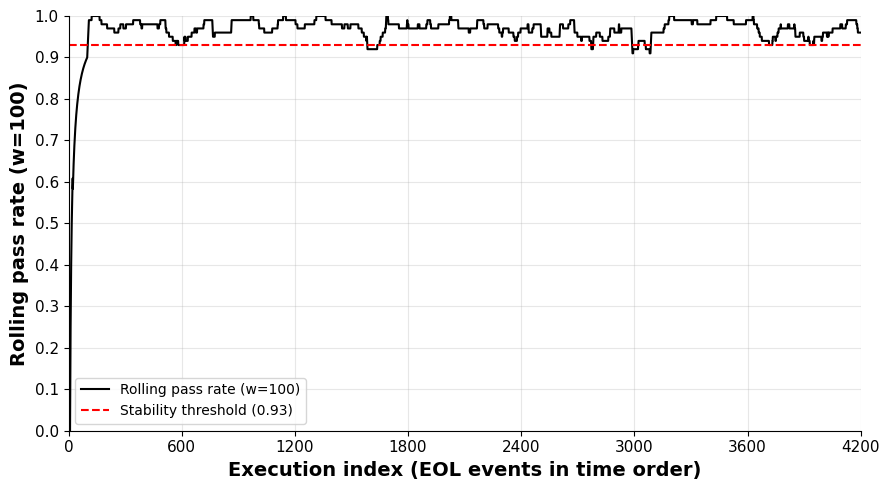}
    \subcaption{EOL validation rolling pass rate ($w=100$), fluctuating around the stability threshold $\tau=0.93$.}
\end{minipage}
\hfill
\begin{minipage}{0.49\textwidth}
    \centering
    \includegraphics[width=\textwidth]{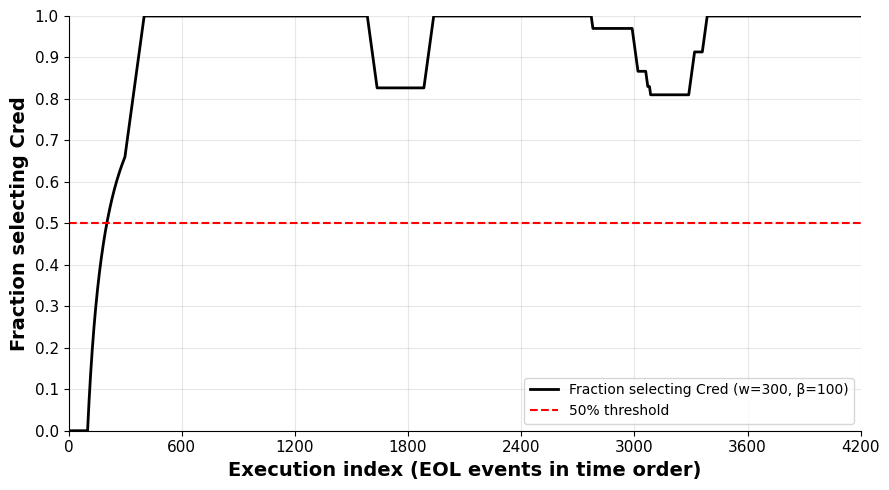}
    \subcaption{Rolling fraction of EOL validation units assigned to $C_{\text{red}}$ by the Thompson Sampling agent ($\beta=100$).}
\end{minipage}
\caption{Rolling pass rate and $C_{\text{red}}$ selection behavior during validation for the FCT and EOL datasets.}
\label{fig:validation_panels}
\end{figure*}

Figure~\ref{fig:validation_panels} summarizes the rolling pass rate and $C_{\text{red}}$ selection behavior during validation for both datasets. Panels (a) and (b) show the FCT validation stream. In contrast to the training period, which contained only one isolated instability event, the validation period is persistently unstable, with the rolling pass rate fluctuating between 0.60 and 0.92 and remaining below the stability threshold $\tau=0.93$. As a result, the agent automatically suppresses $C_{\text{red}}$ selection to below 30\% without retraining, reflecting a conservative response driven entirely by the stability signal. Panels (c) and (d) show the EOL validation stream. Here, the rolling pass rate fluctuates around the same threshold rather than remaining consistently below it, producing a more dynamic behavior: the agent retains high overall $C_{\text{red}}$ selection while still reacting to individual instability events. This enables zero escapes at $\beta=100$ while preserving 87.17\% cost saving.

\section{Discussion}
\label{sec:discussion}
In this study, we proposed an MAB-based model for quality-preserving production test reduction. We evaluate the proposed model on a 2-step test set from a real-world production line of electronics. The results demonstrate that the value of the proposed framework lies not only in reducing test effort, but in doing so while preserving quality under changing production conditions. This is most clearly seen in the validation results reported in Table~\ref{tab:combined_results}. In the FCT stage, the static reduced plan remained highly effective on the training period, achieving 18.78\% test-time saving with zero escapes, but failed under temporal validation with concept drift, producing 110 escaped defects. By contrast, the adaptive Thompson Sampling policy reduced this to 8 escapes at $\beta=10$ and to zero escapes at $\beta=50$, albeit with a corresponding reduction in cost saving. A similar pattern is observed in the EOL stage, where the static reduced plan achieved 91.57\% saving but produced 8 escapes in validation, whereas the adaptive policy achieved zero escapes at $\beta=100$ while still preserving 87.17\% saving. The dynamics underlying these aggregate outcomes are visible in Figure~\ref{fig:validation_panels}, which shows that the policy suppresses selection of $C_{\text{red}}$ when the rolling pass-rate signal indicates instability and resumes reduced testing when conditions recover.

These findings are consistent with the broader shift in manufacturing quality engineering from static inspection design toward closed-loop, data-driven quality management. Recent work on zero-defect manufacturing argues that modern quality systems should combine defect detection, defect prevention, and adaptive decision support across the production chain rather than rely solely on fixed downstream inspection policies~\cite{powell2022advancing}. Likewise, digital quality-management platforms have been proposed to integrate production, process, and quality data in ways that support continuously updated operational decisions rather than isolated offline analyses~\cite{filz2024digitalization}. Viewed in that context, the present framework is not merely a local optimization of testing effort. Rather, it can be interpreted as a practical mechanism for translating unit-level production evidence into adaptive quality-control actions during operation.

The Pareto fronts in Figure~\ref{fig:pareto_panels} make the trade-off of appraisal effort and failure-related cost explicit. In FCT, the frontier shows that a substantial proportion of the achievable saving is concentrated near the zero-escape operating point, indicating that moderate time reduction is possible before defect risk rises sharply. In EOL, the frontier is much steeper and more discrete, reflecting the smaller number of diagnostically relevant steps and the much larger share of non-diagnostic test time. This interpretation is aligned with prior work on multistage inspection planning, which has shown that inspection efficiency depends on the combined effect of process capability, inspection cost, and downstream non-conformance cost~\cite{hamrol2020inspection}. It is also consistent with cost-of-quality analyses demonstrating that the economically preferable inspection strategy is generally the one that balances appraisal savings against the risk of repair, scrap, and escaped failures, rather than the one that simply minimizes inspection time~\cite{farooq2017cost}.

The training-stream behavior shown in Figure~\ref{fig:armselection_panels} further clarifies the role of the adaptive agent. In both stages, the policy learns to favor the reduced plan during stable operating regions, but it does not do so monotonically or irreversibly. Instead, it falls back toward the full plan when the process-stability signal deteriorates, then returns to higher reduced-plan usage when stability is restored. This behavior is important from an industrial standpoint because it suggests that reverting to $C_{\text{full}}$ should not be interpreted as a weakness of the method, but as the intended safety response of a learning policy deployed in a quality-critical environment. In this sense, the framework is closely related to conservative online-learning formulations, in which exploration is permitted only while maintaining performance relative to a trusted baseline policy~\cite{wu2016conservative,kazerouni2017conservative}. In the present application, the full test plan serves as that trusted baseline.

From an application perspective, the results suggest that deployment should proceed gradually and within existing manufacturing quality infrastructures. A practical first step would be shadow-mode deployment, in which the learned policy recommends a plan while the executed plan remains under engineering control. Such a deployment strategy is consistent with recent work on data-driven quality platforms and real-time hybrid inspection systems, both of which emphasize traceability, staged integration, and the use of predictive models to complement rather than abruptly replace established inspection procedures~\cite{filz2024digitalization,ismail2022quality}. In addition, the reward structure of the MAB agent should ultimately be calibrated in plant-specific economic terms, so that the relative utility assigned to test-time reduction, defect detection, rework, scrap, and escape events reflects the actual cost-of-quality structure of the production environment.

This study is not without limitations. First, the framework is evaluated on a single industrial PCBA setting, and broader validation across additional products, production lines, and defect regimes will be required to establish the generality of the observed behavior. Second, the online decision problem is formulated as a binary choice between $C_{\text{full}}$ and a single reduced subset, whereas many production settings may benefit from richer action spaces involving multiple Pareto-optimal subsets or stage-specific test intensities. Third, the reward function compresses multiple operational objectives into a scalar signal, which is convenient for online learning but may under-represent rare, high-consequence escapes. Fourth, the current adaptation mechanism relies on the rolling pass-rate signal as its primary indicator of changing production conditions; although this proved effective in the studied datasets, richer contextual signals from upstream process measurements, material batches, or intermediate quality states may improve responsiveness under subtle forms of drift. Finally, industrial adoption depends not only on performance but also on interpretability, and future work should therefore consider explanation mechanisms that clarify why the policy selected the reduced or full plan for a given unit.

Taken jointly, the results show that adaptive test selection offers a practical middle ground between fully conservative inspection and static test reduction. The principal contribution of the study is therefore not simply that it reduces test time, but that it demonstrates how online learning can be aligned with manufacturing quality objectives in a way that remains operationally cautious under concept drift. By combining offline subset optimization with online policy adaptation, the framework provides a credible path toward more efficient manufacturing test operations without abandoning the central requirement of controlling defect escape.

\section{Discussion}
\label{sec:discussion}
In this study, we proposed an MAB-based model for quality-preserving production test reduction. We evaluate the proposed model on a 2-step test set from a real-world production line of electronics. The results demonstrate that the value of the proposed framework lies not only in reducing test effort, but in doing so while preserving quality under changing production conditions. This is most clearly seen in the validation results reported in Table~\ref{tab:combined_results}. In the FCT stage, the static reduced plan remained highly effective on the training period, achieving 18.78\% test-time saving with zero escapes, but failed under temporal validation with concept drift, producing 110 escaped defects. By contrast, the adaptive Thompson Sampling policy reduced this to 8 escapes at $\beta=10$ and to zero escapes at $\beta=50$, albeit with a corresponding reduction in cost saving. A similar pattern is observed in the EOL stage, where the static reduced plan achieved 91.57\% saving but produced 8 escapes in validation, whereas the adaptive policy achieved zero escapes at $\beta=100$ while still preserving 87.17\% saving. The dynamics underlying these aggregate outcomes are visible in Figure~\ref{fig:validation_panels}, which shows that the policy suppresses selection of $C_{\text{red}}$ when the rolling pass-rate signal indicates instability and resumes reduced testing when conditions recover.

These findings are consistent with the broader shift in manufacturing quality engineering from static inspection design toward closed-loop, data-driven quality management. Recent work on zero-defect manufacturing argues that modern quality systems should combine defect detection, defect prevention, and adaptive decision support across the production chain rather than rely solely on fixed downstream inspection policies~\cite{powell2022advancing}. Likewise, digital quality-management platforms have been proposed to integrate production, process, and quality data in ways that support continuously updated operational decisions rather than isolated offline analyses~\cite{filz2024digitalization}. Viewed in that context, the present framework is not merely a local optimization of testing effort. Rather, it can be interpreted as a practical mechanism for translating unit-level production evidence into adaptive quality-control actions during operation.

The Pareto fronts in Figure~\ref{fig:pareto_panels} make the trade-off of appraisal effort and failure-related cost explicit. In FCT, the frontier shows that a substantial proportion of the achievable saving is concentrated near the zero-escape operating point, indicating that moderate time reduction is possible before defect risk rises sharply. In EOL, the frontier is much steeper and more discrete, reflecting the smaller number of diagnostically relevant steps and the much larger share of non-diagnostic test time. This interpretation is aligned with prior work on multistage inspection planning, which has shown that inspection efficiency depends on the combined effect of process capability, inspection cost, and downstream non-conformance cost~\cite{hamrol2020inspection}. It is also consistent with cost-of-quality analyses demonstrating that the economically preferable inspection strategy is generally the one that balances appraisal savings against the risk of repair, scrap, and escaped failures, rather than the one that simply minimizes inspection time~\cite{farooq2017cost}.

The training-stream behavior shown in Figure~\ref{fig:armselection_panels} further clarifies the role of the adaptive agent. In both stages, the policy learns to favor the reduced plan during stable operating regions, but it does not do so monotonically or irreversibly. Instead, it falls back toward the full plan when the process-stability signal deteriorates, then returns to higher reduced-plan usage when stability is restored. This behavior is important from an industrial standpoint because it suggests that reverting to $C_{\text{full}}$ should not be interpreted as a weakness of the method, but as the intended safety response of a learning policy deployed in a quality-critical environment. In this sense, the framework is closely related to conservative online-learning formulations, in which exploration is permitted only while maintaining performance relative to a trusted baseline policy~\cite{wu2016conservative,kazerouni2017conservative}. In the present application, the full test plan serves as that trusted baseline.

From an application perspective, the results suggest that deployment should proceed gradually and within existing manufacturing quality infrastructures. A practical first step would be shadow-mode deployment, in which the learned policy recommends a plan while the executed plan remains under engineering control. Such a deployment strategy is consistent with recent work on data-driven quality platforms and real-time hybrid inspection systems, both of which emphasize traceability, staged integration, and the use of predictive models to complement rather than abruptly replace established inspection procedures~\cite{filz2024digitalization,ismail2022quality}. In addition, the reward structure of the MAB agent should ultimately be calibrated in plant-specific economic terms, so that the relative utility assigned to test-time reduction, defect detection, rework, scrap, and escape events reflects the actual cost-of-quality structure of the production environment.

This study is not without limitations. First, the framework is evaluated on a single industrial PCBA setting, and broader validation across additional products, production lines, and defect regimes will be required to establish the generality of the observed behavior. Second, the online decision problem is formulated as a binary choice between $C_{\text{full}}$ and a single reduced subset, whereas many production settings may benefit from richer action spaces involving multiple Pareto-optimal subsets or stage-specific test intensities. Third, the reward function compresses multiple operational objectives into a scalar signal, which is convenient for online learning but may under-represent rare, high-consequence escapes. Fourth, the current adaptation mechanism relies on the rolling pass-rate signal as its primary indicator of changing production conditions; although this proved effective in the studied datasets, richer contextual signals from upstream process measurements, material batches, or intermediate quality states may improve responsiveness under subtle forms of drift. Finally, industrial adoption depends not only on performance but also on interpretability, and future work should therefore consider explanation mechanisms that clarify why the policy selected the reduced or full plan for a given unit.

Taken jointly, the results show that adaptive test selection offers a practical middle ground between fully conservative inspection and static test reduction. The principal contribution of the study is therefore not simply that it reduces test time, but that it demonstrates how online learning can be aligned with manufacturing quality objectives in a way that remains operationally cautious under concept drift. By combining offline subset optimization with online policy adaptation, the framework provides a credible path toward more efficient manufacturing test operations without abandoning the central requirement of controlling defect escape.

\section*{Declarations}
\subsection*{Funding}
No funding was received to assist with the preparation of this manuscript.

\subsection*{Conflicts of interest/Competing interests}
The authors have no competing interests to declare that are relevant to the content of this article.

\subsection*{Data availability}
Data is available upon reasonable request from the corresponding author. 

\subsection*{Code availability}
The study's code is freely available in the following GitHub repository: \url{https://github.com/teddy4445/Quality-preserving-Model-for-Electronics-Production-Quality-Tests-Reduction}

\subsection*{Author Contribution}
Noufa Haneefa: Conceptualization, Data Curation, Software, Methodology, Formal analysis, Visualization, Writing – Original Draft.  \\ Teddy Lazebnik: Conceptualization, Formal analysis, Visualization, Investigation, Validation, Supervision, Writing – Original Draft, Writing – Review \& Editing. \\ Einav Peretz-Andersson: Conceptualization, Writing – Review \& Editing, Validation, Supervision. 

\bibliography{example.bib}
\bibliographystyle{unsrt}

\end{document}